\documentclass[runningheads]{llncs}
\usepackage{dingbat}
\usepackage{tabularx}
\usepackage{amsmath}
\usepackage{url}
\usepackage{graphicx}
\usepackage{amssymb}
\usepackage{booktabs}
\usepackage{authblk}
\usepackage{xcolor}

\usepackage[T1]{fontenc}
\usepackage{graphicx}
\title{An AI System for Continuous Knee Osteoarthritis Severity Grading: An Anomaly Detection Inspired Approach with Few Labels}
\author{ Niamh Belton$^{1,2}$ \hspace{0.5cm}  \hspace{0.5cm} Aonghus Lawlor$^{3,4}$ \hspace{0.5cm} Kathleen M. Curran$^{1,2,4}$   
   $^{1}$Science Foundation Ireland Centre for Research Training in Machine Learning 
    $^{2}$School of Medicine, $^{3}$School of Computer Science, University College Dublin 
    $^{4}$Insight Centre for Data Analytics, University College Dublin, Dublin, Ireland 
}
\date{}
\begin{document}
\maketitle

\begin{abstract}

The diagnostic accuracy and subjectivity of existing Knee Osteoarthritis (OA) ordinal grading systems has been a subject of on-going debate and concern.  
Existing automated solutions are trained to emulate these imperfect systems, whilst also being reliant on  large annotated databases for fully-supervised training. This work proposes a three stage approach for automated continuous grading of knee OA that is built upon the principles of Anomaly Detection (AD); learning a robust representation of healthy knee X-rays and grading disease severity based on its distance to the centre of normality. In the first stage, SS-FewSOME is proposed, a self-supervised AD technique that learns the \lq{}normal\rq{} representation, requiring only examples of healthy subjects and $<3\%$ of the labels that existing methods require. In the second stage, this model is used to pseudo label a subset of unlabelled data as \lq{}normal\rq{} or \lq{}anomalous\rq{}, followed by denoising of pseudo labels with CLIP. The final stage involves retraining on labelled and pseudo labelled data using the proposed Dual Centre Representation Learning (DCRL) which learns the centres of two representation spaces; normal and anomalous. Disease severity is then graded based on the distance to the learned centres. The proposed methodology outperforms existing techniques by margins of up to 24\% in terms of OA detection and the disease severity scores correlate with the Kellgren-Lawrence grading system at the same level as human expert performance. Code available at \url{https://github.com/niamhbelton/SS-FewSOME_Disease_Severity_Knee_Osteoarthritis}.

\end{abstract}
\section{Introduction}
Knee Osteoarthritis (OA) is a degenerative joint disease that affects over 250 million of the world's population \cite{kneestats}. Two of the most common grading systems for OA diagnosis are the Kellgren-Lawrence (KL) system \cite{klscale} and the Osteoarthritis Research Society International (OARSI) atlas criteria \cite{oarsi1,oarsi2}. The KL system consists of five ordinal classes from grade zero to four where grade zero is healthy, grade four is severe OA and grades $\ge 2$ are the cut-off for OA diagnosis \cite{compar}. The OARSI atlas criteria diagnoses OA if the subject meets specific criteria relating to the degree of Joint Space Narrowing (JSN) and the severity of osteophytes \cite{oarsi_criteria}. The subjectivity of these scales has been a matter of on-going concern \cite{kl_concern} with studies reporting wide ranges of inter-observer reliability of 0.51 to 0.89 \cite{oa_scales} for the KL system.
Low agreement between the two scales has also been observed, with \cite{compar} reporting that the cut-off for defining knee OA using the two systems should not be considered comparable.
The variability of OA severity grading has been attributed to the difference in the level of clinician's experience and/or the use of subjective language such as \lq{}possible\rq{} osteophytic lipping and \lq{}doubtful\rq{} joint space narrowing in the guidelines for grading \cite{chen}. The challenges of the current OA grading systems suggests that ordinal classes may not be suitable for assessing OA severity and highlights the requirement for an automated system.

This work aims to overcome the weaknesses of current AI systems and models for OA grading.
Firstly, the high performance of existing techniques (accuracies $\ge 90\%$) \cite{image_sharp,3mod} is a cause for concern given the inter-rater reliability between human experts can be as low as 0.51 \cite{oa_scales}, suggesting  that the methods have overfit to the dataset. Secondly, several of the approaches train the model as a multi-class classification problem which do not consider the ordinal nature of the data \cite{lstm,dcaae,multiscale}. 
Thirdly, existing solutions are reliant on large datasets consisting of thousands of X-rays for training, along with ground truth OA severity labels from experts which is a tedious and costly process \cite{inter,siamese,medai-oa}.

This work proposes a three stage approach to designing a continuous automated disease severity system. The principle that underpins the proposed methodology is that healthy knee X-rays can be more easily identified but classifying the degree of OA severity is a subjective process. The first stage therefore, focuses solely on learning a robust representation of healthy X-rays through Self-Supervised Learning (SSL). The model FewSOME \cite{fewsome} is leveraged for this as it requires as few as 30 labels to achieve optimal performance and therefore, the challenge of acquiring a large dataset is eliminated. This work extends FewSOME to use SSL, namely SS-FewSOME. The OA severity of an X-ray is then assessed based on its distance in Representation Space (RS) to the centre of the normal RS, borrowing the core concept of several Anomaly Detection (AD) techniques. The second stage involves using the trained technique to pseudo label unlabelled data and denoising these labels with the CLIP model. The third stage is Dual Centre Representation Learning (DCRL), where the model contrastively learns two representations, normal and anomalous. Disease severity is then graded based on the distance to the learned centres.

\noindent The contributions of this work are;

\begin{itemize}
    \item A newly proposed continuous OA grading system based on the principles of AD that is not trained to overfit to a subjectively graded dataset, thus removing any subjective bias.
    \item We advance the original FewSOME by including SSL, patch based contrastive learning, denoising of pseudo labels with CLIP and DCRL, increasing the AUC for OA detection from 58.3\% (original FewSOME implementation) to 72.9\%.
    \item The proposed three stage method outperforms the existing SOTA on the task of OA detection by 24.3\% (from 56.7\% to 81.0\%).
    \item This performance is achieved using a fraction of the data and less than 3\% of the labels that existing methods require, eliminating the need for extensive computational resources, large annotated datasets and access to expert annotators. Thus, removing the barriers for clinical implementation and improving health equity.

\end{itemize}

\section{Related Work}

\subsection{Automated Knee OA Severity Classification}
Several fully-supervised Convolutional Neural Network (CNN) based approaches have been proposed to classify knee X-rays according to the KL system. Antony et al. \cite{noel} developed a technique to localise the knee joint area on the X-ray and then classify the OA severity using a CNN. The CNN simultaneously minimised a categorical cross-entropy loss and a regression mean squared error loss. Several works improved upon this performance by employing ensembles of CNNs \cite{inter} and implementing more advanced CNNs such as the ResNET \cite{rel_work}, ConvNeXt \cite{novel}, EfficientNetV2-S \cite{eff} and the HRNet coupled with an attention mechanism to filter out the counterproductive features \cite{multiscale}. Siamese networks \cite{siamese} have also been employed by first localising the knee joint area and training an ensemble of classifiers for the lateral and medial components separately. Wahyuningrum et al. \cite{lstm} manually preprocesses the data by cropping the X-rays to the knee joint area. They then use a CNN to extract features, followed by a Long Short-Term Memory model (LSTM) to grade the OA severity. Farooq et al. \cite{dcaae} makes use of both labelled and unlabelled data by first training a dual-channel adversarial autoencoder in an unsupervised manner for the task of reconstruction. They then incorporate two supervised classification networks for KL-grade and leg-side classification achieving an average F1 score of 0.76.  Bose et al. \cite{fss} demonstrated the importance of precise feature selection for knee OA diagnosis. They use a CNN to extract features, followed by feature selection techniques such as Particle Swarm Optimization (PSO) and Genetic Bee Colony (GBC) to identify the most relevant features. These methods were further improved by incorporating complex data pre-processing such as image enhancement \cite{image_sharp}, resulting in an average accuracy of 91\% and noise-reduction with Gaussian-filters, normalising with a pixel-centring method, and implementing a balanced contrast enhancement technique \cite{3mod}. The majority of existing techniques optimised their performance using loss functions suitable for multi-class classification, where the distance between all grades are equidistant i.e. the distance between grade zero and grade four is equal to the distance between grade zero and grade one. Culvenor et al. \cite{chen}, instead, employed an ordinal loss function to take into account the ordinal nature of KL grading system. 

MediAI-OA \cite{medai-oa} has advanced beyond simple KL grade classification by automatically quantifying the degree of JSN in the medial and lateral tibiofemoral joint and detecting osteophytes in the medial distal femur, lateral distal femur, medial proximal tibia and lateral proximal tibia regions. They employ a four step approach. Firstly, they train a RetinaNet \cite{retina} on bounding box annotations to detect the knee joint and regions associated with osteophytes. Secondly, they segment the joint space region by training the HRNet \cite{hrnet} to detect the most medial and lateral corner points of the distal femur and proximal tibia. They then train an image classification network, NASNet \cite{nasnet} to determine the presence of ostephytes. MediAI-OA also trained a NASNet to classify the KL grade.

A prior work investigated the feasibility of a continuous OA system based on Siamese networks, reporting a positive correlation between the KL grading system and the model output \cite{continuous}.  However, despite this progress, the existing solutions are trained to mimic a flawed grading system and they are reliant on large annotated training sets with the MediAI-OA system training on 44,193 labelled radiographs.

\subsection{Anomaly Detection}
AD is a field of study that aims to automatically identify data samples that differ substantially from normality. AD algorithms train on a dataset, $X_{train}$ which consists of solely normal images. A network $\phi$ learns to extract the features of normality from the training set through various methods such as one-class, generative or Self-Supervised techniques. Given, a test instance $x_t$, the trained network $\phi$ can identify if it belongs to the normal or anomalous class.  One-class methods such as DeepSVDD \cite{deepsvdd} and PatchCore \cite{patchcore} use $\phi$ to learn a RS for normality. They can then identify anomalies based on their distance to normal RS, where large distances indicate that the case is an anomaly and small distances indicate that the case is normal.  
Generative models are another class of AD models \cite{riad}, particularly popular in the medical field \cite{ddad,miccaigan,proxy}. At training time, they learn to reconstruct the image. At test time, they can then identify anomalies by large reconstruction errors. However, SSL based methods such as Cutpaste \cite{cutpaste} have been dominating the field of AD in recent years, with several works leveraging SSL for medical AD tasks \cite{salad,ddad}. Although the majority of AD techniques train on unlabelled data, a recent technique Dual-distribution discrepancy with self-supervised refinement (DDAD) \cite{ddad} trains on both labelled and unlabelled data. They train an ensemble of reconstruction networks with the objective of modelling the normative distribution and a second ensemble of networks to model the unknown distribution. They then calculate the intra-discrepancy and inter-discrepancy of the distributions to assign anomaly scores. Finally they train a separate network via self-supervised learning to further refine the anomaly scores. They demonstrated their AD performance on chest X-rays and brain MRI.

There has been recent interest in Few Shot AD for settings where there is limited data. Few Shot AD focuses on learning to detect anomalies having trained on zero shots or few shots of the normal class \cite{winclip,fewsome}. FewSOME \cite{fewsome} is a recent few shot AD technique originally developed for poor quality data detection in medical imaging datasets \cite{semi} and has since proven its performance for AD in natural images, industrial defect detection and motion artefact detection in brain MRI \cite{mad}. FewSOME is in the category of one-class AD models as it contrastively learns the normal RS from solely nominal images and prevents representational collapse with the use of an \lq{}anchor\rq{} and stop loss. Anomalies are identified based on their distance to the normal RS.

\section{Materials}
\begin{figure}[t!]  \includegraphics[width=\linewidth]{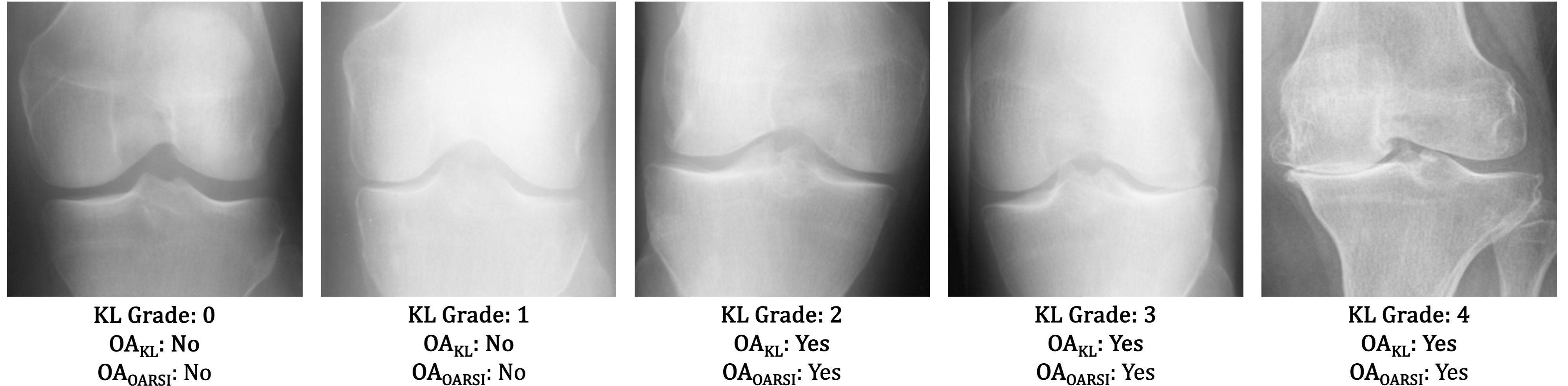}
  \caption{Examples of X-rays from the training set for each KL grade. OA$_{KL}$ refers to whether OA was diagnosed based on the KL scale and OA$_{OARSI}$ refers to whether OA was diagnosed based on the OARSI criteria.}

  \label{fig:oa_example}
\end{figure}

The baseline cohort from the Osteoarthritis Initiative (OAI) \cite{dataset,oai} was used for this analysis as it is the most commonly used subset in the literature. The OAI is an observational, longitudinal study of 4,796 subjects across 431,000 clinical studies with the goal of better understanding of prevention and treatment of knee osteoarthritis \cite{oai}. The images are of size $224\times 224$, with $5,778$ available for training, $826$ available for validation and $1,526$ available for testing. It was ensured that both knees belonging to a single patient were within the same dataset split to prevent any data leakage. The dataset provides the KL grades and a grading for both the severity of JSN and osteophytes on a scale from zero to three. The grading for JSN and osteophytes are provided for medial and lateral tibiofemoral compartments separately. According to the OARSI scale, OA is diagnosed if any of the following criteria are met; either JSN grade $\ge2$, sum of osteophyte grades $\ge2$ or grade one JSN in combination with a grade one osteophyte \cite{compar,oarsi_criteria}. The dataset contained missing values for osteophyte gradings in some cases. As the missing values solely occurred when the KL grades were equal to zero or one and the JSN for both the medial and lateral compartments had a grade of zero, we labelled these cases as not having OA according to the OARSI scale. Examples of a knee X-ray for each grade are shown in figure \ref{fig:oa_example}. Table \ref{tab:oa_dataset} shows the breakdown of the class balances for each split. 

\begin{table}[!htbp]
    \caption{The number of X-rays in each dataset split and the number of X-rays that were diagnosed with OA according to the KL system (OA$_{KL}$) and the OARSI system (OA$_{OARSI}$).}
    \begin{center}
        \begin{tabularx}{\textwidth}{l|XXXXXXX}
            \hline \hline
            Split & Grade 0 & Grade 1 & Grade 2 & Grade 3 & Grade 4 & OA$_{KL}$ & OA$_{OARSI}$ \\
            \hline
           Train & 2286 & 1046 & 1516 & 757 & 173 & 2446 & 2436\\
Validation  & 328 & 153 & 212 & 106 & 27 & 345 & 342\\
Test & 639 & 296 & 447 & 223 & 51 & 721 & 706\\
        \end{tabularx}
    \end{center}
    \label{tab:oa_dataset}
\end{table}

\vspace{-1.1cm}

\section{Methods}

\vspace{-0.1cm}

\subsection{Stage 1: Self-Supervised Learning with FewSOME (SS-FewSOME)}

\begin{figure}[hbt!]  \includegraphics[width=\linewidth]{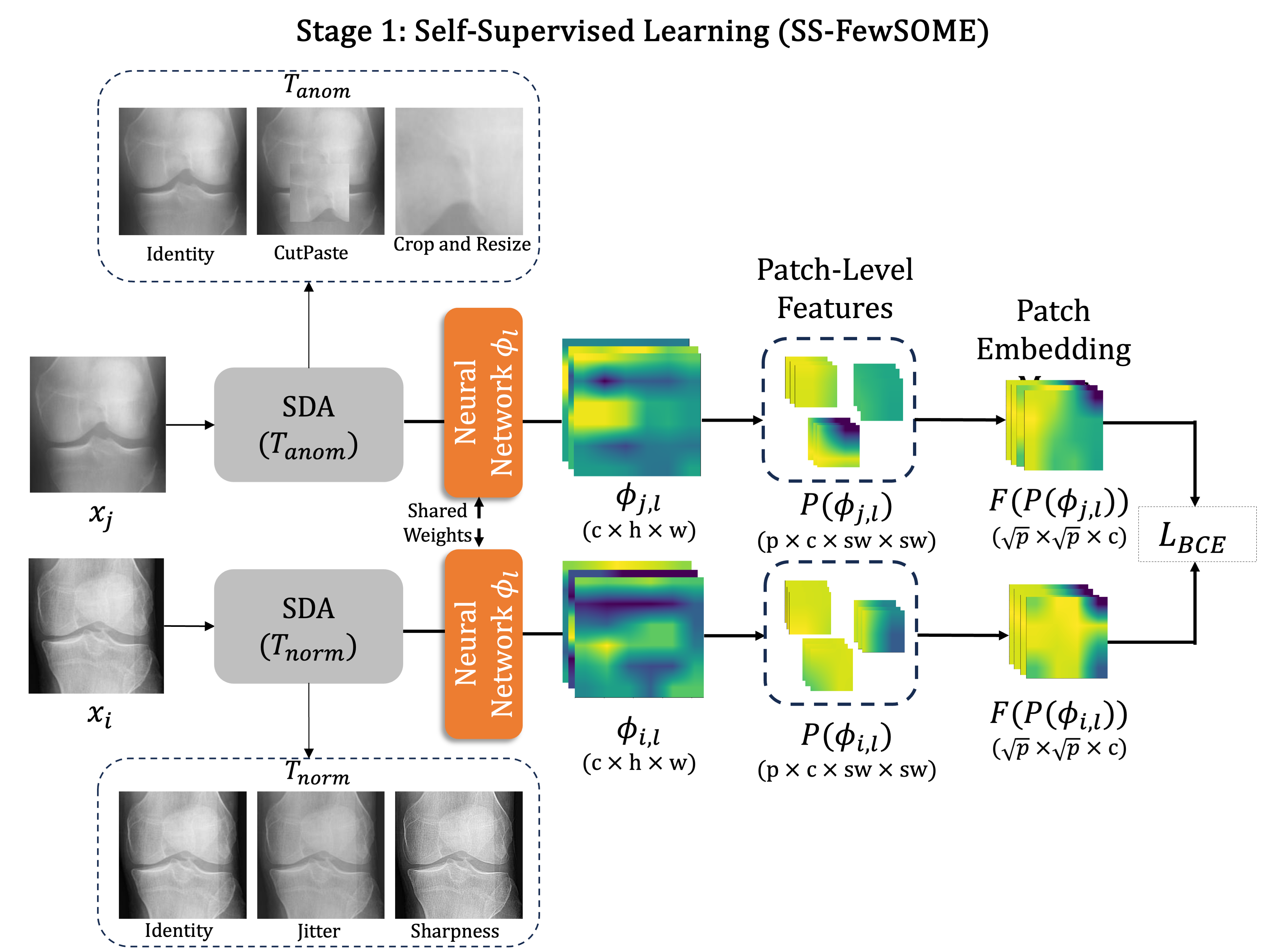}
  \caption{The figure shows iteration $i$ of training where a data sample $x_{i} \in X_N$ is paired with a randomly selected data sample $x_j \in X_N$. Following the SDA, both instances are input in tandem into the neural network $\phi_l$. The output, denoted as $\phi_{i,l}$ and $\phi_{j,l}$ is converted to patch level features, $P(\phi_{i,l})$ and $P(\phi_{j,l})$. Following average pooling, the tensor is reshaped to create patch embedding maps,  $F(P(\phi_{i,l}))$ and $F(P(\phi_{j,l}))$. The patch embedding map, $F(P(\phi_{i,l}))$ consists of $p$ patches at coordinates $z, k$, where each patch is denoted as $p_{i,k,z}$ The Cosine Distance (CD) is calculated between each patch, $p_{i,z,k}$ and $p_{j,z,k}$ at the same coordinates $z,k$. The loss is calculated as the Binary Cross Entropy between the CD between patches and each ground truth label, $y_{i,x,k}$}
\vspace{-0.3cm}

  \label{fig:oa_stage1}
\end{figure}

\vspace{-0.1cm}

The purpose of Stage 1, Self-Supervised FewSOME (SS-FewSOME) is to learn a representation of healthy X-rays. There are $K$ ensembles trained on a dataset $X_{N}$ of size $N$, where $X_N$ is sampled from the training data, $X_{train}$ which consists of solely nominal images of healthy X-rays.
In iteration $i$ of training, a data sample $x_{i} \in X_N$ is paired with a randomly selected data sample $x_j \in X_N$. The original FewSOME implementation relies on ImageNet pretrained weights and an \lq{}anchor\rq{} for learning compact representations of the normal class where the ground truth label, $y$ is equal to zero for the duration of training. This work extends this method to use SSL where both data samples, $x_i$ and $x_j$ are input into a Stochastic Data Augmentation (SDA) module. The SDA module consists of two sets of transforms, $T_{norm}$ and $T_{anom}$. The set of transforms $T_{norm}$ are applied only to $x_i$ and they consist of weak, global augmentations that aim to generate additional X-rays whose representation are within the hyper-sphere of the  normal RS. $T_{norm}$ consists of the identity function (i.e. no transformation), applying jitter to the image and adjusting the sharpness and brightness. The set of transforms $T_{anom}$ are applied only to $x_j$ and they consist of the identity function and strong augmentations, random cropping followed by resizing and Cutpaste \cite{cutpaste}. Section \ref{sec:oa_tran_anom} presents the results of experiments conducted to choose the optimal performing transforms for $T_{anom}$.

Following the SDA, the first $l$ layers of a neural network, $\phi$, initialised with ImageNET \cite{imagenet} pre-trained weights are used to transform the input space, $X_{N}$ to the RS, $\phi_{l}(X_{N}) \in \mathbb{R}^{c \times h \times w}$, where $c$ is the number of channels, $h$ is the height and $w$ is the width of the extracted feature maps. The later layers are removed as they are biased towards natural image classification. As the size of the training dataset is small, both $\phi_l(x_i)$ and $\phi_l(x_j)$, denoted as $\phi_{i, l}$ and $\phi_{j, l}$, are converted to patch level features,  $P(\phi_{i, l})$ and $P(\phi_{j, l})$ using a sliding window of size $sw \times sw$ with $stride =1$ with resulting dimensions of $p \times c \times sw \times sw$ where $p$ is the number of patches. The patch level features also improve the sensitivity of the model to localised anomalies such as osteophytes. Feature aggregation is performed by average pooling for each patch and the resultant tensor is reshaped to create patch embedding maps, $F(P(\phi_{i,l}))$ and $F(P(\phi_{j,l})) \in \mathbb{R}^{\sqrt{p} \times \sqrt{p} \times c}$. The patch embedding map, $F(P(\phi_{i,l}))$ consists of $p$ patches at coordinates $z, k$, where each patch is denoted as $p_{i,z,k}$. 
 
This method exploits that X-rays are spatially similar and therefore, the Cosine Distance, CD is calculated only between patches $p_{i,z,k} \in F(P(\phi_{i,l}))$ and $p_{j,z,k} \in F(P(\phi_{j,l}))$ that have the same coordinates, $z,k$. The ground truth label $y_{i,z,k} \forall (z,k)$ is equal to zero if the identity transform in the SDA was applied to $x_j$ and $y_{i,z,k} \forall (z,k)$ is equal to one if cropping and resizing was applied. In the case of applying the CutPaste transform, the entire image is not affected and therefore,  $y_{i,z,k}$ is equal to one if the receptive field of $p_{j, z, k}$ was affected by the CutPasting and zero otherwise. The CD between each patch, $p_{i,z,k}$ and $p_{j,z,k}$ is calculated as shown in equation \ref{eq:oa_pred} and denoted as $\hat{y}_{i,z,k}$. The difference between $\hat{y}_{i,z,k}$ and the ground truth label, $y_{i,z,k}$ are minimised using Binary Cross Entropy during training as shown in equation \ref{eq:oa_loss}. Stage 1 is depicted in figure \ref{fig:oa_stage1}.

\begin{equation}
\label{eq:oa_pred}
\hat{y}_{i,z,k} = 1 - \frac{p_{i, z,k} \cdot p_{j, z, k}}{|| p_{i, z,k} || ||  p_{j, z,k} ||}
\end{equation}

\begin{equation}
\label{eq:oa_loss}
\mathcal{L}_{BCE} = -\frac{1}{N \cdot \sqrt{p} \cdot \sqrt{p}}  \sum_{i=1}^{N} \sum_{z=1}^{\sqrt{p}}  \sum_{k=1}^{\sqrt{p}}  y_{i,z,k} \cdot log(\hat{y}_{i,z,k}) \\
+ (1-y_{i,z,k}) \cdot log(1-\hat{y}_{i,z,k})
\end{equation}

\subsection{Stage 2: Pseudo Labelling and Denoising with CLIP}\label{sec:oa_stage2}

\begin{figure}[hbt!]  \includegraphics[width=\linewidth]{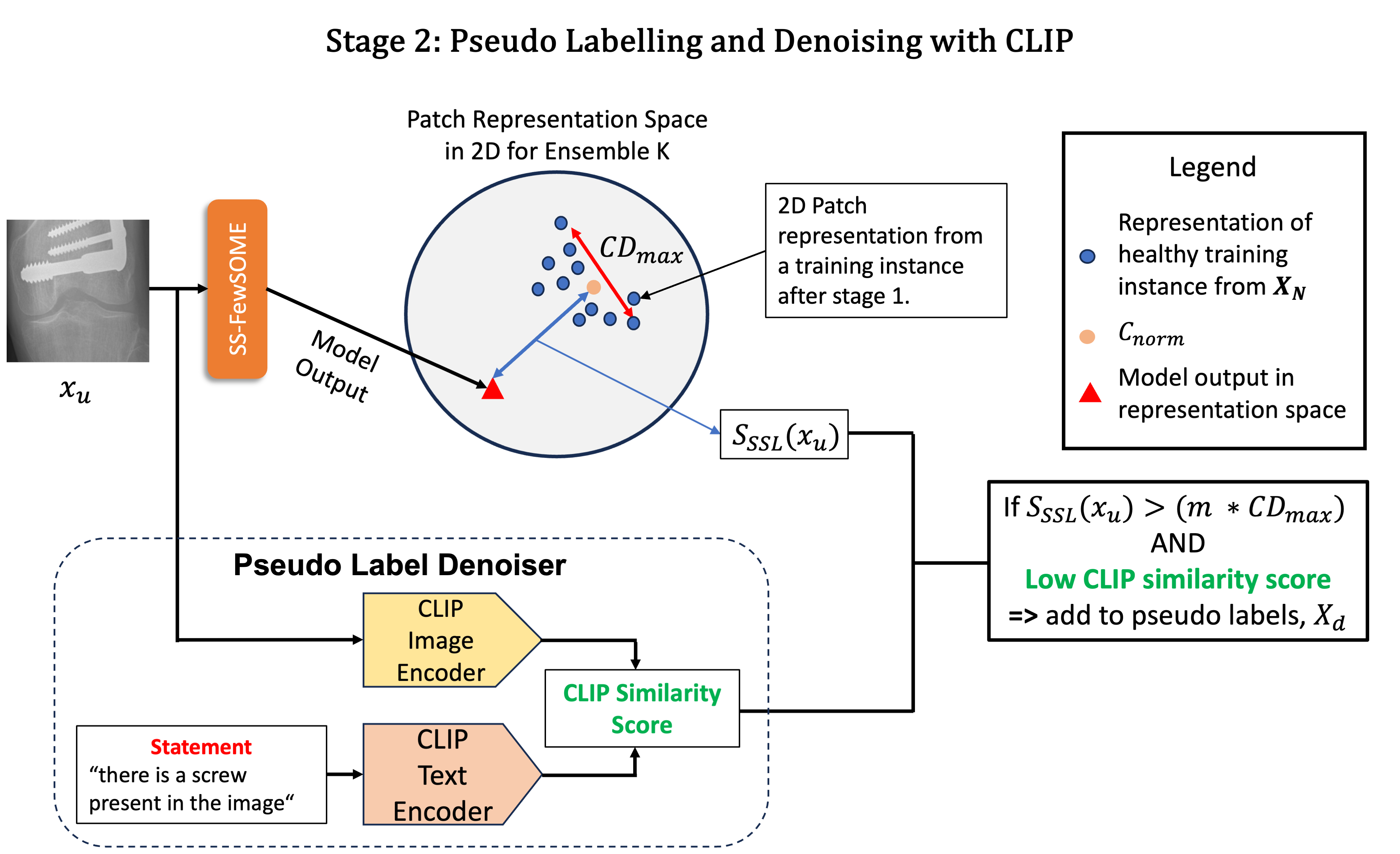}
  \caption{SS-FewSOME from Stage 1 is used to transform $x_u \in X_u$ to the RS. The AS scoring function, $S_{SSL}$ is then used to pseudo label $x_u$. The sample $x_u$ is also input into the CLIP image encoder and its similarity is calculated between the image encoding and the text encoding of the compositional statement. If $S_{SSL}(x_u) > (m \times CD_{max}$), $x_u$ is added to the pseudo labels, $X_d$ for Stage 3. }

  \label{fig:oa_stage2}
\end{figure}

Given an unlabelled dataset of X-rays, $X_u$ of size $u$ where $X_u \cap X_N = \emptyset$, the patch embedding map, $F(P(\phi_{u,l}))$ consisting of patches $p$, for each data sample $x \in X_u$ is obtained. The centre, $C_{norm}$ of the normal RS is then calculated as in equation \ref{eq:oa_c}. An anomaly scoring function, $S_{SSL}$ assigns $x_u$ an Anomaly Score (AS) equal to the average CD to the centre $C_{norm}$, as shown in equation \ref{eq:oa_s}.

\begin{equation}
\label{eq:oa_c}
C_{norm} = \frac{1}{N \cdot \sqrt{p} \cdot \sqrt{p}}  \sum_{i=1}^{N} \sum_{z=1}^{\sqrt{p}}  \sum_{k=1}^{\sqrt{p}} p_{i,z,k}
\end{equation}

\begin{equation}
\label{eq:oa_s}
S_{SSL}(x_u) = \frac{1}{ \sqrt{p} \cdot \sqrt{p}}   \sum_{z=1}^{\sqrt{p}}  \sum_{k=1}^{\sqrt{p}} 1 - \frac{p_{u, z,k} \cdot C_{norm}}{|| p_{u, z,k} || ||  C_{norm} ||}
\end{equation}

Each ensemble $K$ identifies a data instance as an anomaly if the AS is greater than $m$ times the maximum CD between all possible pairs in the training data (denoted as $CD_{max}$), where $1 \le m \le \infty$. Data instances that have been voted as an anomaly across all ensembles are then assumed to be severe OA cases. The model can be retrained on these cases to improve the sensitivity to OA specific anomalies with the newly assigned pseudo label of $y=1$. However, knee X-rays often contain anomalies unrelated to the OA grade, such as the presence of screws and metal in the X-ray due to knee replacement implants \cite{metal}. Using the pre-trained CLIP \cite{clip} model with ViT-B/32 Transformer architecture, the similarity between each image and a single compositional statement,\lq{}there is a screw present in the image\rq{}, was obtained. This statement was chosen based on the analysis in section \ref{sec:oa_compositional}. The cosine similarity was calculated between the compositional statement and all X-rays in the validation data. To identify a threshold, $t_c$ for detecting screws in X-rays, the lowest similarity score of the subset in the validation set that contained a screw was found, resulting in $t_c=23.05$. In Stage 2, X-rays with CLIP similarity scores $> t_c$ were identified as \lq{}false\rq{} anomalies. These are removed from the pseudo labels, resulting in a dataset $X_d$ of denoised pseudo labels. This stage is conducted only during training. No X-rays are removed from the dataset at test time.

\subsection{Stage 3: Dual Centre Representation Learning with FewSOME}

\begin{figure}[hbt!]  \includegraphics[width=\linewidth]{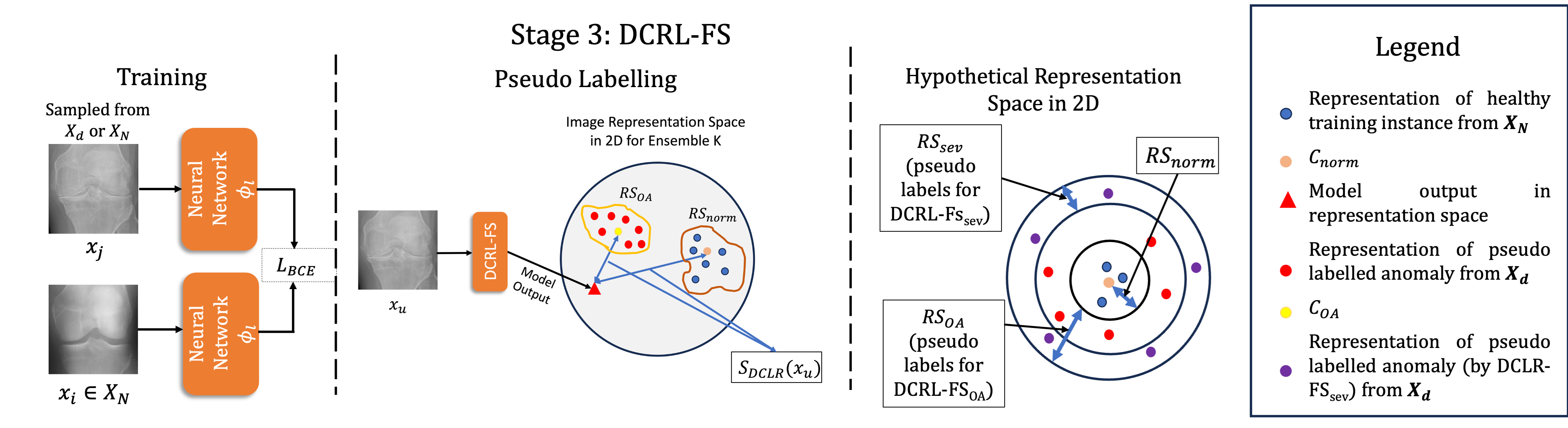}
  \caption{Stage 3 uses DCRL to learn two representation spaces, one for normal instances, $RS_{norm}$ and another for anomalies $RS_{OA}$ (or $RS_{sev}$). The figure also shows a hypothetical representation space demonstrating where the pseudo labels are obtained for DCRL-FS$_{sev}$ and DCRL-FS$_{OA}$. }

  \label{fig:oa_stage3}
\end{figure}

Given there are now anomalous pseudo labels available for training, this section proposes Dual Centre Representation Learning (DCRL) with FewSOME. The goal of DCRL is to simultaneously learn two separate RSs for normality and anomalies. DCRL has the same methodology as Stage 1 except the SDA is switched off and the value of $y$ is determined by where $x_j$ is sampled from. For example, in iteration $i$ of training, $x_i \in X_N$ is paired with a randomly selected data sample $x_j$ where $y_i=0$ if $x_j \in X_N$ and $y_i=1$ if $x_j \in X_d$. The objective of the training is to transform $x_i$ and $x_j$ to the RS, calculate the CD between the two instances and compare this to the ground truth label $y_i$. 

It was found that the model learns a more robust representation of healthy X-rays when training with pseudo labels of varying OA severity, while it is less robust when exposed to only pseudo labels of severe OA cases (cases with high AS scores output from the model). However, training the model on pseudo labels of more severe OA cases (cases with high AS) results in a highly accurate severe OA detector but a less accurate OA detector. This trade-off can be combatted by training two models, DCRL-FS$_{sev}$ and DCRL-FS$_{OA}$. 

DCRL-FS$_{OA}$ is the continuous disease severity grading system that learns a RS for healthy cases, $RS_{norm}$ and another for all other varying severity OA cases, $RS_{OA}$. DCRL-FS$_{sev}$ is used to learn a RS for healthy cases, $RS_{norm}$ and another for severe OA cases, $RS_{sev}$ and behaves as a severe OA detector. The margin, $m$, described in section \ref{sec:oa_stage2}, controls the level of severity of the pseudo labels. Therefore $m$ is set higher when training DCRL-FS$_{sev}$ and lower when training DCRL-FS$_{OA}$. DCRL-FS$_{OA}$ can be trained iteratively, meaning it can be used to predict additional pseudo labels, which are denoised using the method of Stage 2 and then retrained on the newly assigned pseudo labels.

The models can then be combined to improve the OA and severe OA detection accuracy. This model is denoted as DCRL-FS$_{comb}$. The CD to the centre $C_{norm}$ of $RS_{norm}$ (obtained from training DCRL-FS$_{sev}$) is denoted as $d_1$ and the CD to $C_{sev}$ of $RS_{sev}$ is denoted as $d_2$. The anomaly scoring function, $S_{DCRL_{sev}}$ is then equal to the difference between the distances $d_1$ and $d_2$ as can be seen from equations \ref{eq:oa_d1} to \ref{eq:oa_sdcrl}. Similarly, $S_{DCRL_{OA}}$ is calculated as in equation \ref{eq:oa_sdcrloa}, where $d_3$ is the CD to the centre $C_{norm}$ of $RS_{norm}$ (obtained from training DCRL-FS$_{OA}$) and $d_4$ is the CD to the centre $C_{OA}$ of $RS_{OA}$. Given the range of possible output scores for $d1$, $d2$, $d3$ and $d4$ is $[0,2]$, the range of possible output values from $S_{DCRL_{OA}}$ and $S_{DCRL_{sev}}$ is $[-2,2]$. These scores are normalised to be between zero and one by adding the minimum value of the range, two and adding the inter-range value of four. The DCRL-FS$_{sev}$ behaves as a severe OA detector, meaning if $S_{DCRL_{sev}}$ is greater than a threshold, $t$, $x_u$ is assigned a score of $1 + S_{DCRL_{sev}}(x_u)$, as $0 \le S_{DCRL_{OA}}(x_u) \le 1$. Otherwise it is equal to $S_{DCRL_{OA}}(x_u)$, as shown in equation 9.

\begin{equation}
\label{eq:oa_d1}
d_1 =  1 - \frac{\textit{DCRL-FS}_{sev}(x_u) \cdot C_{norm}}{|| \textit{DCRL-FS}_{sev}(x_u) || ||  C_{norm} ||}
\end{equation}

\begin{equation}
\label{eq:d2}
d_2 =  1 - \frac{\textit{DCRL-FS}_{sev}(x_u) \cdot C_{sev}}{|| \textit{DCRL-FS}_{sev}(x_u) || ||  C_{sev} ||}
\end{equation}

\begin{equation}
\label{eq:oa_sdcrl}
S_{DCRL_{sev}} = \frac{(d_1 - d_2) + 2}{4}
\end{equation}

\begin{equation}
\label{eq:oa_sdcrloa}
S_{DCRL_{\mathrm{OA}}} = \frac{(d_3 - d_4) + 2}{4}
\end{equation}

\[
    S_{DCRL_{comb}}(x_u)= 
\begin{cases}
     1 + S_{DCRL_{sev}}(x_u),& \text{if } S_{DCRL_{sev}}(x_u) > t\\
    S_{DCRL_{OA}}(x_u),              & \text{otherwise}
\end{cases} \hspace{2.3cm}(9)
\]

\begin{figure}[t!]  \includegraphics[width=\linewidth]{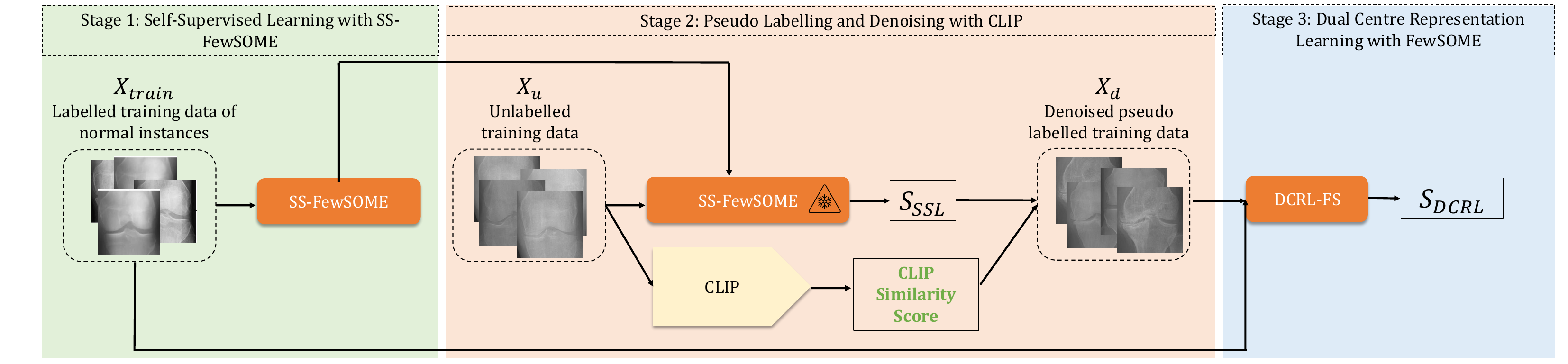}
  \caption{ simplified overview of the proposed three stage approach.}

  \label{fig:oa_overall}
\end{figure}

The output of $S_{DCRL_{OA}}$ is continuous with values in the range $[0,1]$ after normalisation. The AS of the combined model, DCRL-FS$_{comb}$, designed to improve OA and severe OA detection has a discontinuity in its output values when combining $S_{DCRL_{OA}}$ and $S_{DCRL_{sev}}$ using the method in equation 9. When combining the models for $DCRL-FS_{comb}$, the scores between zero and one should be interpreted as representing severity levels from healthy to moderate, while scores $>1$ should be interpreted as representing the level of severity within severe cases. Despite the discontinuity, the ranking of scores is preserved, meaning that scores can be interpreted relative to each other. Figure \ref{fig:oa_overall} presents an overview of the three stage approach.

\section{Implementation Details}
The implementation details are summarised in table \ref{tab:oa_summary}.

\textbf{Hyper-Parameters.} The hyper-parameters and model backbones that had the best performance on the validation set were selected. The models were trained with $K=10$, learning rate of $1e-06$, batch size of one, weight decay of $0.1$ and the Adam optimiser \cite{adam}. The first five layers of the AlexNET \cite{alexnet} was used as the model backbone for Stage 1, while the larger VGG-16 \cite{vgg} was implemented for later stages where more training data was available. As previously mentioned, DCRL-FS$_{OA}$ can be trained iteratively, meaning it can be used to predict additional pseudo labels, which are denoised using the method of Stage 2 and then retrained on the newly assigned pseudo labels. There was no increase in the performance of DCRL-FS$_{OA}$ after two iterations on the validation set. The value of $t$ is set based on the distribution of output scores from DCRL-FS$_{sev}$ on the complete training set. Using the anomaly score in the 95\% percentile resulted in the optimal performance on the validation set and therefore, this method for defining the threshold was implemented.

\textbf{Margin, $m$.} \hspace{0.02cm} The code implementation of this method automatically selects the margin value of $m$ based on the number of desired pseudo labels. The optimal performing number of pseudo labels on the validation set was $N=30$ for DCRL-FS$_{OA}$ and $N=3$ for DCRL-FS$_{sev}$, resulting in margin values of 1.184 and 3.122 respectively. Keeping the margin static at 1.184 for the remainder of training resulted in 263 pseudo labelled anomalies in the second iteration of Stage 3 DCRL-FS$_{OA}$.

\textbf{Training Data Size.} \hspace{0.02cm} Table \ref{tab:oa_summary} summarises the training data that was available at each stage and the hyper-parameters. Based on the analysis shown in section \ref{sec:oa_rep}, the labelled training set size, $X_{train}$ was equal to 150 examples. $K=10$ ensembles were trained with $N=30$ for each ensemble as per the original FewSOME implementation. 
In the second/final iteration of Stage 3, DCRL-FS$_{OA}$, there was substantially more data available for training due to the 263 pseudo labelled anomalies. To avoid a large class imbalance, the number of training instances for each ensemble was increased from $30$ to $150$ (all available training data).

\textbf{Early Stopping.} \hspace{0.02cm} For Stage 1, FewSOME's early stopping method of halting training once the training loss began to plateau was employed. For all implementations in Stage 3, the training was stopped when the CD between the normal centre, $C_{norm}$ and anomalous centres $C_{OA}$ (when training DCRL-FS$_{OA}$) or $C_{sev}$ (when training DCRL-FS$_{sev}$) begins to plateau. Section \ref{sec:oa_es} presents an analysis of the effects of this early stopping technique.

\begin{table*}
\caption[Number of Training Examples Available at each Stage and Hyper-Parameters]{The number of training examples available at each stage and hyper-parameters. Stage 3 is abbreviated to S3.}
\label{tab:oa_summary}

\begin{center}
\begin{tabular}{l|ccccc}
\hline \hline
\textbf{Stage} & \textbf{\# Labelled} & \textbf{\# Pseudo} & \textbf{Architecture} & \textbf{K} & \textbf{m} \\
 & \textbf{Normal} & \textbf{Labelled} & & & \\
 & \textbf{Samples} & \textbf{Anomalies} & & & \\
\hline 
Stage 1 & 150 & 0 & AlexNet\textsubscript{l=5} & 10 & $\times$ \\
S3, DCRL-FS\textsubscript{sev} & 150 & 3 & VGG-16 & 10 & 3.122 \\
S3\textsubscript{iter1}, DCRL-FS\textsubscript{OA} & 150 & 30 & VGG-16 & 10 & 1.184 \\
S3\textsubscript{iterfinal}, DCRL-FS\textsubscript{OA} & 150 & 263 & VGG-16 & 10 & 1.184 \\
\hline
\end{tabular}
\end{center}
\end{table*}

\subsection{Competing Methods}

\subsubsection{Stage 1}
SS-FewSOME's performance is compared to competing methods, DeepSVDD \cite{deepsvdd}, SOTA AD technique PatchCore \cite{patchcore} and few shot AD technique FewSOME \cite{fewsome}. As the original implementation of DeepSVDD is compatible with small images of 32$\times$32, the images are downsampled to 128$\times$128 and the architecture is upscaled by increasing the number of kernels. Competing methods were trained on $150$ samples of normal X-rays as was SS-FewSOME.

\subsubsection{Stage 3 Final Iteration}
The final iteration of the model is compared to DDAD \cite{ddad}, a SOTA model in medical AD, as it trains on both labelled and unlabelled data. DDAD is trained on $150$ labelled healthy knee X-rays and 5,628 unlabelled X-rays. This is the most similar set-up to the stage 3 of the proposed method as it trains on 150 labelled healthy knee X-rays and there were 5,628 unlabelled X-rays available for pseudo labelling.

\section{Evaluation}
The methodology is assessed based on its ability to diagnose OA according to the KL scale (grades $\ge 2$) and the OARSI scale. The Area Under the Receiver Operating Characteristic Curve (AUC) is used to evaluate both tasks. The AUC is calculated by comparing the output of $S_{SSL}$ with the ground truth label for Stage 1. It is calculated by comparing the output of $S_{DCRL_{OA}}$, $S_{DRCRL_{sev}}$ and $S_{DRCRL_{comb}}$ with the ground truth label for Stage 3. The Spearman Rank Correlation Coefficient (SRC) between the model output and the KL grade is also reported. Finally, to ensure that the model can identify severe OA cases to a high degree of accuracy, the performance of the models on detecting cases with KL grades $> 3$ is also calculated. 


\section{Results}
Table \ref{tab:oa_res} reports the AUC in \% for the task of OA detection according to the KL and OARSI grading systems, denoted as $AUC_{KL}$ and $AUC_{O}$ respectively. The SRC between the model output and the KL grade is also reported as $SRC_{KL}$. The table also reports the performance of detecting cases with KL grades $> 3$ as $AUC_{KL_{g>3}}$. The results were averaged across five seeds and one standard deviation is reported. 



 \begin{table}
\caption{AUC in \% with one standard deviation over five seeds. The standard deviations of ensemble methods are the standard deviations of performance between each ensemble. SS-FewSOME abbreviated to SS-FS. The $*$ highlights the two models that were combined for DCRL-FS$_comb$ in stage 3$_{iter1}$}
\begin{center}
\resizebox{\columnwidth}{!}{%
\begin{tabular}{lcc|cccc}
\hline \hline
\
Method & Stage & Patches & AUC$_{KL}$ & AUC$_{O}$   & AUC$_{KL_{g}>3}$ & SRC$_{KL}$\\
\hline 
        DeepSVDD \cite{deepsvdd} & 1 &  - &  $54.8\pm{}1.0$ & $55.8\pm{}0.9$  & $62.3\pm{}2.2$ & $0.097\pm{}0.0$  \\
    Patchcore \cite{patchcore}& 1  &  - &$64.6\pm{}0.6$ & $65.7\pm{}0.5$  & $89.1\pm{}0.6$ & $0.270\pm{}0.0$ \\

    FewSOME \cite{fewsome}& 1&  \checkmark  &  $58.3\pm{}0.2$  & $54.8\pm{}3.4$ & $67.8\pm{}0.7$ & $0.144\pm{}0.5$\\
    SS-FS  & 1&  $\times$ &   $66.8\pm{}0.7$  & $66.8\pm{}0.9$ & $85.4\pm{}0.4$ & $0.315\pm{}0.0$\\

    SS-FS  & 1  & \checkmark &  $\mathbf{72.9\pm{}1.0}$  & $\mathbf{73.7\pm{}1.1}$ & $\mathbf{89.4\pm{}1.9}$ & $\mathbf{0.432\pm{}0.0}$ \\
    \hline

        DCRL-FS$_{OA}^*$ &3$_{iter 1}$ &   $\times$  &  $78.6\pm{}1.4$  & $79.5\pm{}1.2$ & $95.2\pm{}1.5$ & $0.539\pm{}0.0$ \\ 

        DCRL-FS$_{sev}^*$ &3 &  $\times$ &  $76.9\pm{}1.1$  & $77.3\pm{}1.1$ & $97.6\pm{}0.4$ & $0.491\pm{}0.0$ \\ 
        DCRL-FS$_{comb}^{**}$ &3$_{iter 1}$ &   $\times$ &  $\mathbf{78.6\pm{}1.4}$  & $\mathbf{79.5\pm{}1.2}$ & $\mathbf{97.2\pm{}0.4}$ & $\mathbf{0.541\pm{}0.0}$ \\

        \hline
    DCRL-FS$_{OA}$  & 3$_{iter final}$ &   $\times $&  $81.0\pm{}0.2$  & $81.6\pm{}0.1$ & $95.7\pm{}0.4$ & $0.580\pm{}0.0$ \\ 

    DCRL-FS$_{comb}$ & 3$_{iter final}$ &   $\times$ &  $\mathbf{81.0\pm{}1.4}$  & $\mathbf{81.6\pm{}1.2}$ & $\mathbf{97.6\pm{}0.4}$ & $\mathbf{0.583\pm{}0.0}$ \\

 DDAD  \cite{ddad} & -   &  - &  $56.7\pm{}0.6$ & $57.2\pm{}0.6$  & $69.7\pm{}0.9$ & $0.130\pm{}0.0$ \\
    
\end{tabular}
}
\end{center}
  
\label{tab:oa_res}

\end{table}

\subsection{Stage 1}

Each method had $150$ training instances available for training. The proposed SS-FewSOME significantly outperforms competing methods and a paired t-Test shows a statistically significant difference between the performance of the original FewSOME implementation and the proposed, increasing the $AUC_{KL}$ from $58.3$\% to $72.9$\%. The results also show that integrating the patches method into the model has resulted in a performance improvement, increasing $SRC_{KL}$ from 0.315 to 0.432. 

\begin{figure}[b!]  \includegraphics[width=\linewidth]{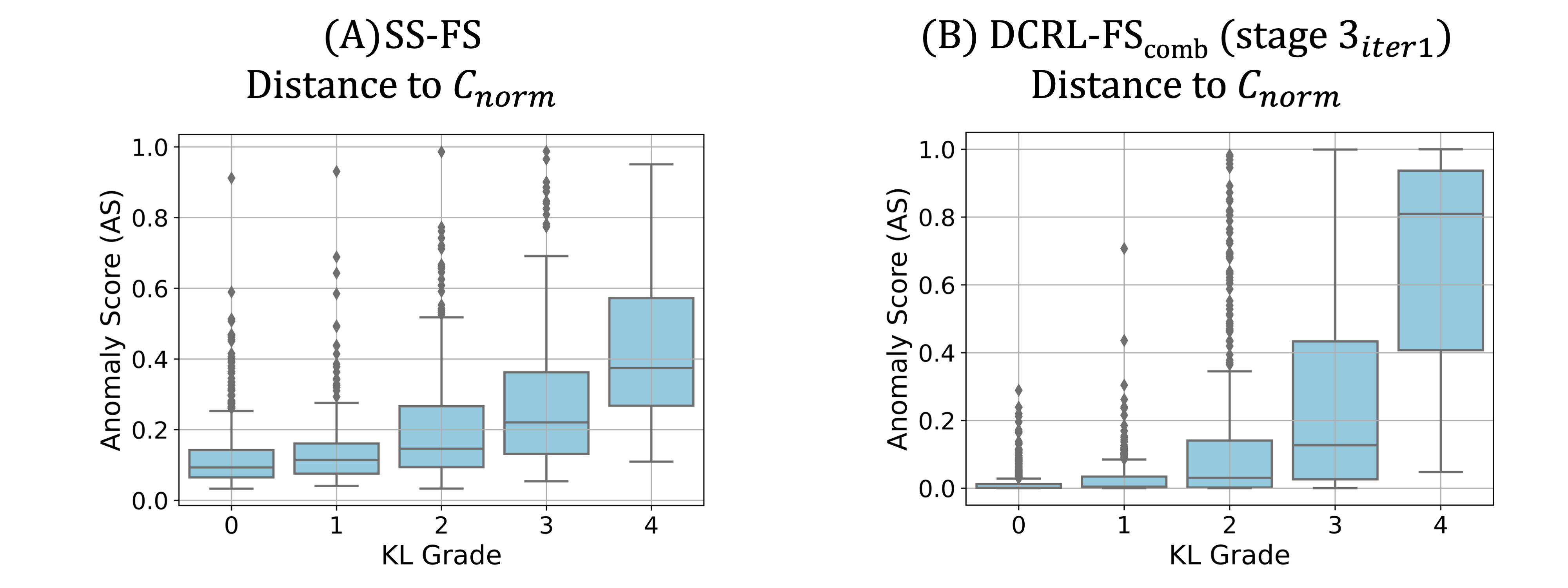}
  \caption{Boxplot of distances of each instance from the test set to $C_{norm}$ in RS at (A) stage 1, SS-FewSOME (SS-FS) and (B) stage 3$_iter1$}
  \label{fig:oa_box}

\end{figure}

\subsection{Stage 3$_{iter1}$}
\begin{figure}[t!]  \includegraphics[width=\linewidth]{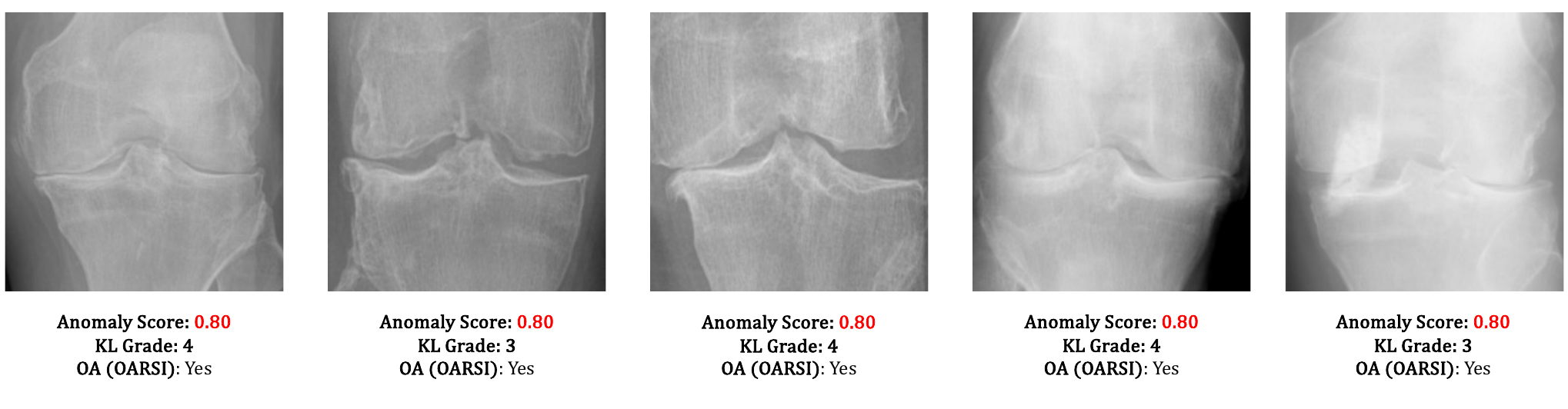}
  \caption{Examples that were assigned the highest anomaly scores by DCRL-FS$_{OA}$ (stage $3_{iter1}$). These examples are severe OA cases and display OA characteristics such as JSN and osteophytes. }

  \label{fig:oa_severe}
\end{figure}

The continuous disease severity grading system, DCRL-FS$_{OA}$ improves the SRC from 0.432 in Stage 1 to 0.539 in Stage 3$_{iter1}$. 
The trade-off between OA detection and severe OA detection accuracy can be evidently seen from the results of DCRL-FS$_{OA}$ (stage $3_{iter1}$) and DCRL-FS$_{sev}$, with the former outperforming DCRL-FS$_{sev}$ in terms of OA detection but underperforming in terms of severe OA detection. However, the combined model DCRL-FS$_{comb}$ (stage $3_{iter1}$) combines both aspects of the models to accurately detect OA, severe OA and correlate with the KL grades with an SRC of 0.541. Figure \ref{fig:oa_box} shows the distance of each data instance in the test set from $C_{norm}$ at stage 1 and stage 3$_{iter1}$. Separation between classes can be evidently seen at stage 1 and furthermore at stage 3$_{iter1}$. Figure \ref{fig:oa_severe} presents the most anomalous examples as output by DCRL-FS$_{comb}$ (stage $3_{iter1}$). The common symptoms of OA are apparent such as JSN and the presence of osteophytes.

\subsection{Stage 3$_{iterfinal}$}
The final iteration achieved an SRC of 0.583, a promising result given the inter-observer reliability of human experts was reported to be as low as 0.51 in some cases \cite{oa_scales}. The proposed methodology outperforms the competing method DDAD \cite{ddad} by margins of up to 24\%, which was trained on $150$ labelled healthy knee X-rays and 5,628 unlabelled X-rays. A paired t-Test showed this to be a statistically significant difference in performance. Additionally, a Two-Sample t-Test showed that the difference between the mean model output for each KL grade was statistically significant at the 5\% significance level.

\vspace{-0.1cm}

\section{Further Analysis}
\vspace{-0.1cm}
\subsection{CLIP Analysis}
\subsubsection{Compositional Statement Ablation}\label{sec:oa_compositional} 
This analysis presents the metal detection performance of CLIP on the validation data for variations of compositional statements. The similarity score between each statement and the image in the validation set was calculated. The AUC was calculated between the output similarity scores and a label that indicates whether there was metal present in the image or not. The statement "there is a screw present in the image" resulted in the optimal performance with 99.3\%.

\begin{table}
\caption[Comparison of the Performance of CLIP on Metal Detection for Variations of Compositional Statements.]{Comparison of the performance of CLIP on metal detection for variations of compositional statements. }
\begin{center}
\begin{tabular}{lc}
\hline \hline
\
Compositional Statement & AUC (\%)\\
\hline 
        "there is a screw present in the image" & 99.3 \\
        "screw" & 98.5 \\ 
        "there is a metal present in the image" & 90.3 \\
        "metal" & 94.0 \\
        "there is another object in the image that is not a knee X-ray" & 2.0 \\

\end{tabular}
 \end{center}
  
\label{tab:oa_clip}
\end{table} 

\subsubsection{CLIP Ablation} \vspace{0.2cm}

\begin{figure}[!t]
\centering 
\includegraphics[width=80mm]{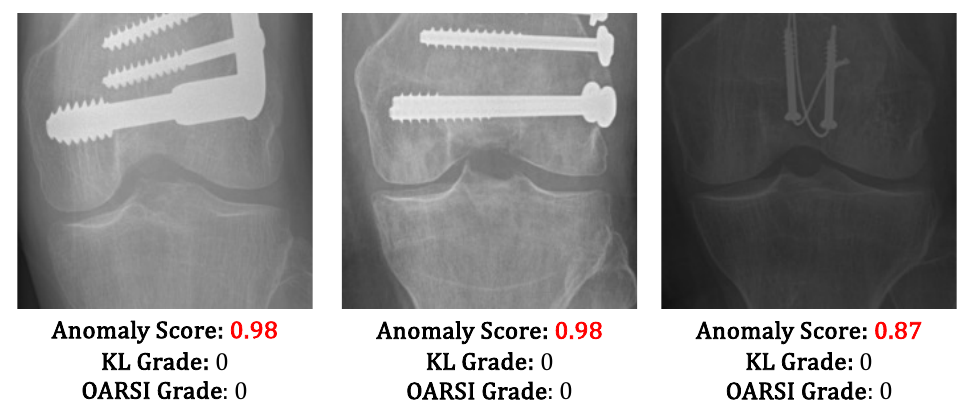}
  \caption{Samples from $X_u$ that were assigned high anomaly scores from SS-FewSOME. These examples were then denoised using the CLIP model and they were removed from the pseudo labels for Stage 3, DCRL training. }

  \label{fig:oa_clip}
\end{figure}

To demonstrate the effectiveness of denoising the pseudo labels with CLIP, DCRL-FS$_{OA}$ at stage 3$_{iter1}$ was also trained without denoising. This resulted in a performance degradation as can be seen from table \ref{tab:oa_clip}. Figure \ref{fig:oa_clip} presents examples of false anomalies that were denoised by CLIP.

\begin{table}
\caption{AUC in \% with one standard deviation over five seeds. }
\begin{center}
\resizebox{\columnwidth}{!}{%
\begin{tabular}{lc|cccc}
\hline \hline
\
Method & Stage & AUC$_{KL}$ & AUC$_{O}$   & AUC$_{KL_{g}>3}$ & SRC$_{KL}$\\
\hline 
       
        DCRL-FS$_{OA}$ &3$_{iter 1}$ &    $\mathbf{78.6\pm{}1.4}$  & $\mathbf{79.5\pm{}1.2}$ & $\mathbf{95.2\pm{}1.5}$ & $\mathbf{0.539\pm{}0.0}$ \\

      DCRL-FS$_{OA}$ (\textbf{No\ CLIP}) &3$_{iter 1}$ &   $75.6\pm{}2.0$  & $76.3\pm{}1.9$ & $93.6\pm{}1.1$ & $0.485\pm{}0.0$ \\

\end{tabular}
}
 \end{center}
  
\label{tab:oa_clip}
\end{table}

\subsection{Analysis of Transforms for $T_{anom}$}
\label{sec:oa_tran_anom}

\begin{figure}[!htbp]  \includegraphics[width=\linewidth]{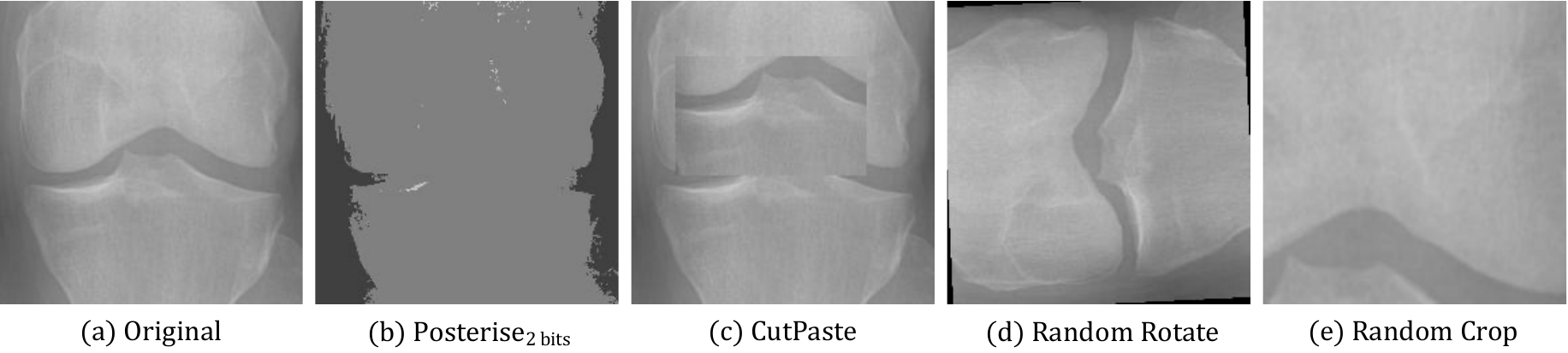}
  \caption{Examples of transforms that were assessed for inclusion in $T_{anom}$ for Stage 1 SSL.}
  \label{fig:oa_ss}
\end{figure}

\vspace{-0.2cm}
The following transforms were assessed for inclusion in $T_{anom}$; posterising which limits the number of tones and colours in the image, random rotate, random crop and CutPaste \cite{cutpaste} which takes a segment of the image and pastes it elsewhere on the image. Table \ref{tab:oa_tanom} presents the results of each transform on the validation set. CutPaste outperforms the other transforms by a substantial margin. As a combination of cropping and CutPaste resulted in the optimal performance, they were selected for the final model. Examples of the transforms are shown in figure \ref{fig:oa_ss}.


 \begin{table}
\centering
   \caption{The table shows the performance of each transform from $T_{anom}$ on the validation set.}

\begin{tabular}{l|ccccc}
\hline \hline
\
Method & AUC$_{KL}$ & AUC$_{O}$   & AUC$_{KL_{g}\ge3}$ & SRC$_{KL}$\\

     \hline
 Posterise$_{2 bits}$ & $59.8\pm{}1.5$ &$60.2\pm{}1.7$  & $67.6\pm{}2.1$ & $0.174\pm{}0.0$  \\
     Rotate$_{>90^{\circ},<270^{\circ}}$  & $58.4\pm{}0.8$  & $58.0\pm{}0.7$ &  $77.2\pm{}3.4$  & $0.140\pm{}0.0$\\
     Crop & $59.6\pm{}1.0$ & $59.6\pm{}1.0$ &  $76.7\pm{}1.5$  & $0.177\pm{}0.0$\\
    CutPaste & $72.1\pm{}1.1$  &$72.4\pm{}1.3$&$87.1\pm{}2.4$& $0.388\pm{}0.0$\\
     Crop + CutPaste& $\mathbf{73.3\pm{}0.7}$ & $\mathbf{72.9\pm{}0.7}$ &  $\mathbf{92.9\pm{}1.0}$  & $\mathbf{0.400\pm{}0.0}$

  \end{tabular}

  \label{tab:oa_tanom}

\end{table}

\subsection{Training Set Size Analysis} \label{sec:oa_rep}

\begin{figure}[!htbp]
\centering
\includegraphics[width=70mm]{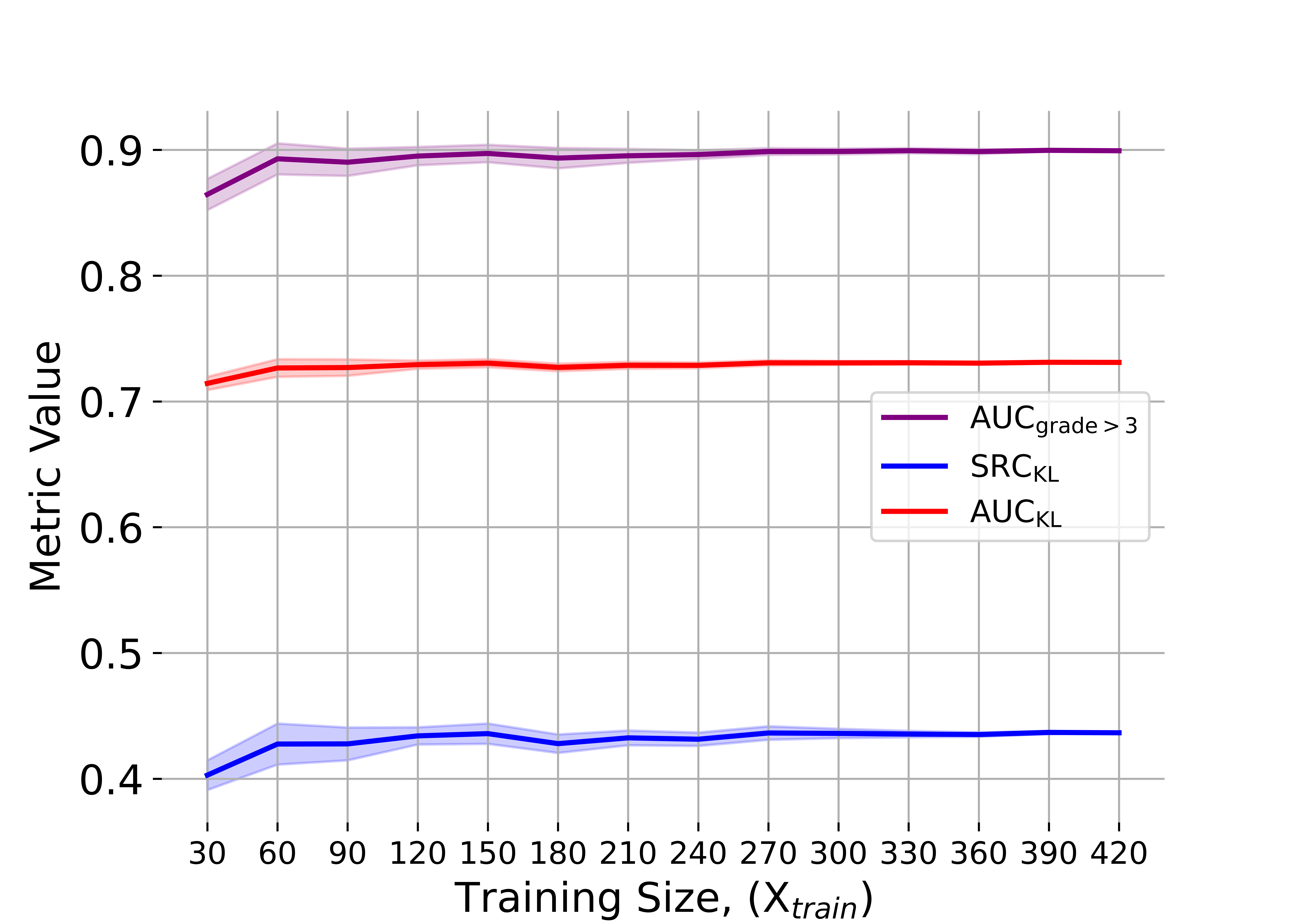}
  \caption{The average result for each metric, on the test set for stage 1 for varying training set sizes, $X_{train}$.  }

  \label{fig:oa_rep}

\end{figure}

Figure \ref{fig:oa_rep} shows the model performance for each metric for varying sizes of $X_{train}$. As SS-FewSOME in stage 1 trains on $N=30$, each training set size represents an ensemble of models training on $N=30$. For example, a training set size of $60$ is two ensembles training on $N=30$. It can be seen from the figure that there a plateau in performance begins between $90$ to $150$ training instances. Therefore, the experiments were conducted training on $150$ labelled instances. The figure also demonstrates how SS-FewSOME can achieve optimal performance on small training set sizes of as low as 90 examples.

\subsection{Analysis of Early Stopping} \label{sec:oa_es}
\begin{figure}[!htbp]  \includegraphics[width=\linewidth]{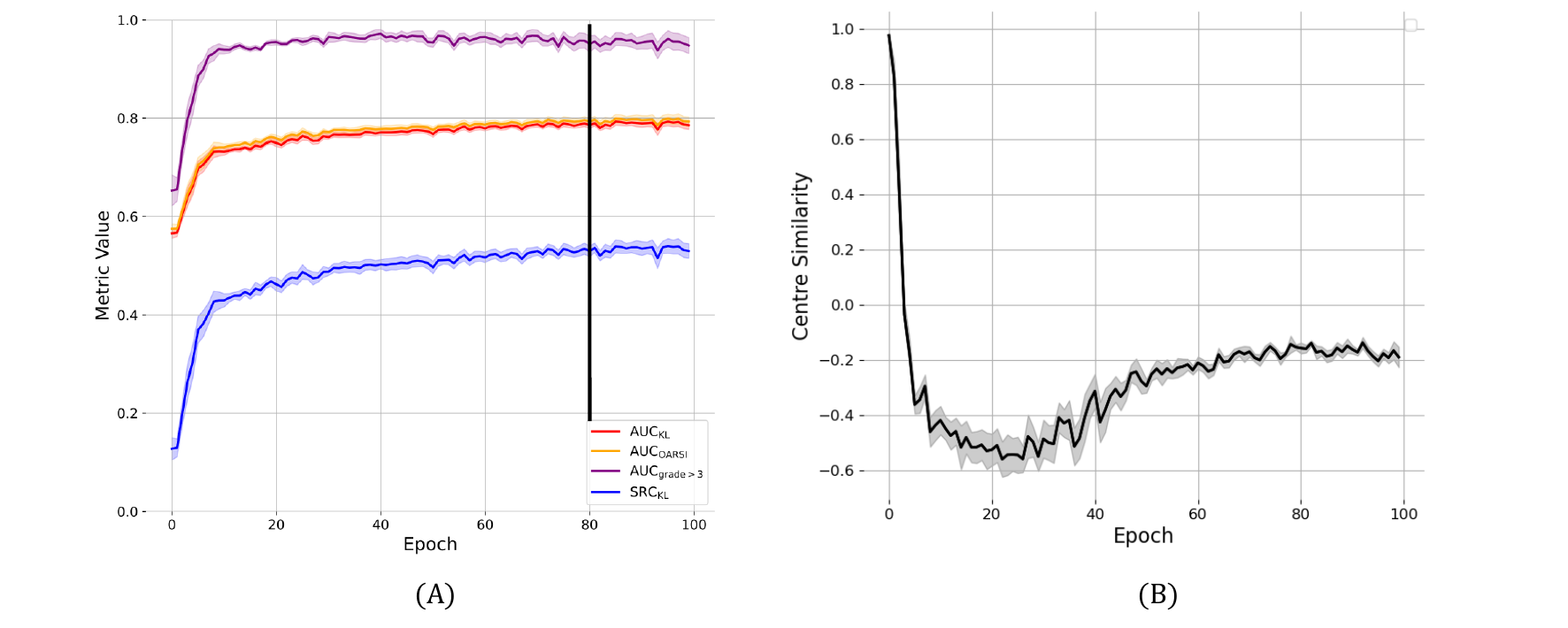}
  \caption{(A) shows the average value for each metric ($AUC_{KL}$, $AUC_{OARSI}$, $AUC_{grade>3}$ and $SRC_{KL}$) over ten seeds for each epoch on the test set for Stage 3$_iter1$. The black vertical line shows where the average early stopping occurred over the ten seeds. Early stopping is based on the plateau of the centre similarities as seen in (B).}

  \label{fig:oa_es}
\end{figure}

In stage 3, training is halted when the CD between the centres, $C_{norm}$ and $C_{OA}$ begins to plateau. Figure \ref{fig:oa_es} (A) shows the average value for each metric ($AUC_{KL}$, $AUC_{OARSI}$, $AUC_{grade>3}$ and $SRC_{KL}$) over ten seeds for each epoch on the test set for DCRL-FS$_{OA}$ at stage $3_{iter1}$. The black vertical line shows where early stopping occurred on average. It can be seen from the figure that there is no further performance increase after early stopping and overfitting has not occurred by the time of early stopping. Figure \ref{fig:oa_es} (B) shows the CD between centres $C_{norm}$ and $C_{OA}$ at each epoch. 

\section{Discussion and Conclusion}

This work has presented a three stage approach to developing a continuous disease severity grading system. The output anomaly scores of the proposed approach have a moderate to strong correlation with OA disease severity, achieving an SRC value of 0.58. Additionally, the proposed model can detect OA with an AUC of 0.81 according to both the KL and OARSI scales. The severe OA detection performance of the continuous grading system, DCRL-FS$_{OA}$ can be improved from 95.7\% to 97.6\% by combining it with the severe OA detector DCRL-FS$_{sev}$, to form DCRL-FS$_{comb}$. Although this causes a discontinuity in scores, the ranking of severity is preserved. The analysis showed that the proposed method outperforms existing methods by significant margins and can achieve human level performance whilst training on few labels.

The advantages of this approach over existing approaches fully supervised approaches are three fold. Firstly, the proposed method eliminates the requirement for substantial amounts of labelled data, reducing the barriers to clinical implementation. Secondly, this method is not trained to mimic a subjective grading system. Thirdly, diagnosing on a continuous grading system holds promise for substantial clinical impact by enabling the assessment of disease severity at any stage along a continuum, rather than predefined discrete categories. This method is particularly useful for capturing subtle variations and changes over time, providing a more detailed and dynamic understanding of the condition's progression, ultimately resulting in improved patient care.

A limitation of this work is that its evaluation relies on comparisons with the KL grading system and the OARSI system, both of which are subjectively graded. The gold standard for evaluating the proposed system is assessment by clinicians. One potential method of evaluation would involve relative comparisons, where clinicians are presented with two knee X-rays and asked to determine which case has more severe OA. Their rankings could then be compared to the rankings generated by the proposed system. Additionally, clinicians could be asked to express their preference between the discrete KL grading system and the proposed continuous grading system for example cases. This would provide a high quality evaluation of the proposed system. An additional limitation is that the method requires the fine tuning of various hyper-parameters such as $m$ to labelled validation data. 

Future work could explore additional datasets such as MOST \cite{most} to further assess the generalisability of the approach. Additionally, the translation of such a system into clinical practice would require further work on explainability. Techniques such as Grad-CAM \cite{gradcam} could be employed to visualize the most salient regions of an image, providing insight into the system's decision-making process. Additionally, a more comprehensive analysis would be needed to interpret what the values on the proposed scale represent in terms of OA features, such as the degree of JSN and the severity of osteophytes. Techniques such as Testing with Concept Activation Vectors (TCAV) \cite{tcav} could be employed to quantify the importance of various visual biomarkers of OA in knee X-rays, thus improving the interpretability of the system.

 \section*{Acknowledgements}
This work was funded by Science Foundation Ireland through the SFI Centre for Research Training in Machine Learning (Grant No. 18/CRT/6183). This work is supported by the Insight Centre for Data Analytics under Grant Number SFI/12/RC/2289 P2.

\bibliographystyle{splncs04}
\bibliography{main.bib}

\section*{Appendix}
\subsection*{Selection of $m$}
This section presents the results on the validation set obtained from training on varying quantities of pseudo labelled training instances ($d$) at Stage 3 DCRL-FS$_{OA}$. Table \ref{tab:oa_margin} shows in the AUC in \% on the task of OA detection according to the KL grading system for values of $d$ ranging from 10 to 40. It can be observed from the table that the performance begins to plateau at $d=30$. As $d=30$ results in a class balance with the number of labelled training cases of healthy examples, $N=30$, $d$ was set to 30, resulting in margin $m$ value of 1.184. Table \ref{tab:oa_margin2} shows the AUC in \% for the task of severe OA detection at Stage 3 DCRL-FS$_{sev}$. It can be observed that $d=3$ is the best performing implementation on the validation set, resulting in a value of $m$ of 3.122.

\begin{table}[!htbp]
    \caption{AUC in \% on the task of OA detection according to the KL grading system at Stage 3 DCRL-FS$_{OA}$.}
    \begin{center}
        \begin{tabular}{>{\centering\arraybackslash}p{3cm}|>{\centering\arraybackslash}p{4cm}}
            \hline \hline
            $d$ &  DCRL-FS$_{OA}$\\
            \hline
           10 & 77.8 \\ 
           15 & 77.7\\ 
           20 & 77.7 \\
           25 & 77.8 \\ 
           30 & \textbf{78.0 }\\ 
           35 & \textbf{78.0} \\ 
           40 & \textbf{78.0}
        \end{tabular}
    \end{center}
    \label{tab:oa_margin}
\end{table}

\begin{table}[!htbp]
    \caption{AUC in \% on the task of severe OA detection according to the KL grading system at Stage 3 DCRL-FS$_{sev}$.}
    \begin{center}
        \begin{tabular}{>{\centering\arraybackslash}p{3cm}|>{\centering\arraybackslash}p{4cm}}
            \hline \hline
            $d$ &  DCRL-FS$_{sev}$\\
            \hline
           1 & 90.5 \\ 
           2 & 96.6\\ 
           3 & \textbf{97.4} \\
           4 & 97.1 \\ 
           5 & 97.3\\ 
           
        \end{tabular}
    \end{center}
    \label{tab:oa_margin2}
\end{table}

\end{document}